# Collaborative Learning-Enhanced Lightweight Models for Predicting Arterial Blood Pressure Waveform in a Large-scale Perioperative Dataset

Wentao Li[†]
*College of Health Science and Environmental Engineering*
*Shenzhen Technology University*
Shenzhen, China

Yonghu He[†]
*College of Health Science and Environmental Engineering*
*Shenzhen Technology University*
Shenzhen, China

ZiRong Yu
*College of Health Science and Environmental Engineering*
*Shenzhen Technology University*
Shenzhen, China

Kun Gao
*Department of Critical Care Medicine*
*Jinling Hospital*
Suzhou, China

Qing Liu*
*School of Advanced Technology*
*Xi'an Jiaotong-Liverpool University*
*Suzhou, China*
*Email: Qing.Liu@xjtlu.edu.cn*

Yali Zheng*
*College of Health Science and Environmental Engineering*
*Shenzhen Technology University*
*Shenzhen, China*
*Email: zhengyali@sztu.edu.cn*

***Abstract*—Noninvasive arterial blood pressure (ABP) monitoring is essential for patient management in critical care and perioperative settings, providing continuous assessment of cardiovascular hemodynamics with minimal risks. Numerous deep learning models have developed to reconstruct ABP waveform from noninvasively acquired physiological signals such as electrocardiogram and photoplethysmogram. However, limited research has addressed the issue of model performance and computational load for deployment on embedded systems. The study introduces a lightweight sInvResUNet, along with a collaborative learning scheme named KDCL_sInvResUNet. With only 0.89 million parameters and a computational load of 0.02 GFLOPS, real-time ABP estimation was successfully achieved on embedded devices with an inference time of 8.49 milliseconds for a 10-second output. We performed subject-independent validation in a large-scale and heterogeneous perioperative dataset containing 1,257,141 data segments from 2,154 patients, with a wide BP range (41-257 mmHg for SBP, and 31-234 mmHg for DBP). The proposed KDCL_sInvResUNet achieved comparable performance to large models, with a mean absolute error of 10.06 mmHg and mean Pearson correlation of 0.88 in tracking ABP changes. Despite these promising results, all deep learning models showed significant performance variations across different demographic and cardiovascular conditions, highlighting their limited ability to generalize across such a broad and diverse population. This study lays a foundation work for real-time, unobtrusive ABP monitoring in real-world perioperative settings, providing baseline for future advancements in this area.**

## I. INTRODUCTION

In critical care and perioperative settings, blood pressure (BP) may fluctuate rapidly, and sustained hypotension can lead to insufficient perfusion of vital organs, increasing the risk of postoperative complications involving the heart, brain, and kidneys[1], [2]. Continuous arterial blood pressure (ABP) monitoring provides rich information on cardiovascular dynamics beyond intermittent cuff-based measurements, enabling more effective hemodynamic management during surgery and intensive care [3], [4].

Despite its clinical value, continuous ABP monitoring is commonly achieved using invasive arterial lines, which are associated with risks such as bleeding, thrombosis, and infection, and are therefore limited to high-risk patients [5]. In addition, certain populations, including infants and patients with arteriosclerosis, are not ideal candidates for invasive catheterization. In contrast, cuff-based noninvasive BP devices only provide intermittent measurements and may cause discomfort during repeated inflation. These limitations have motivated the development of cuffless and noninvasive approaches for continuous ABP waveform estimation [6], [7].

Compared with hardware-dependent methods such as volume clamp and arterial tonometry, which require specialized sensors and complex instrumentation, reconstructing ABP waveforms from readily available physiological signals, including electrocardiogram (ECG) and photoplethysmogram (PPG), offers a cost-effective and easily deployable solution. Recent studies have demonstrated that deep learning models have potential to reconstruct ABP waveforms from ECG and PPG signals using various architectures, including convolutional and recurrent neural networks [8], [9], [10], encoder–decoder frameworks [11], [12], and transformer-based models [13], [14], [15].

Although these approaches have achieved promising accuracy, most existing models are computationally intensive and involve large parameter sizes, which significantly limits their deployment on resource-constrained embedded platforms

This work was supported by Young Scientists Fund from National Natural Science Foundation of China (NSFC) (62301333), XJTLU Research Development Fund (RDF-21-02-068), and National Science Foundation of China (NSFC) (62373259). *Corresponding authors. [†] Equally contributed.

commonly used in bedside monitoring systems. Only a limited number of studies [14], [16] have explored embedded implementation of ABP reconstruction models, and their performance has rarely been validated on large-scale datasets using subject-independent evaluation protocols.

In this work, we propose a lightweight one-dimensional Inverted Residual U-Net (InvResUNet) architecture for real-time ABP waveform reconstruction from ECG and PPG signals. The proposed model is specifically designed for embedded deployment by leveraging inverted residual blocks and depthwise separable convolutions to achieve high parameter efficiency. Furthermore, a collaborative distillation strategy is introduced to improve performance of the lightweight model without increasing inference complexity. The proposed method is evaluated on a large perioperative dataset using subject-independent validation and is deployed on embedded platforms to demonstrate real-time performance.

## II. Methods

### A. Model structure

A novel lightweight model, i.e., 1-D Inverted Residual UNet (InvResUNet), is proposed for real-time ABP reconstruction on embedded devices by using a combination of building blocks, as shown in Fig. 1. The model is modified based on the U-Net architecture, and the encoder adopted the inverted residual block of MobileNetV2 [17]. The details of these building blocks are described as follows.

The inverted residual block, as shown in Fig. 1(a), initially expands the number of input channels through pointwise convolution operations to extract features across channels, then extracts channel-independent features via depthwise convolutions, and finally projects the expanded features to the original dimension through another set of pointwise convolutions. Additionally, it adds the residual connections acting as linear bottleneck, to allow gradients flowing directly through multiple layers, preventing nonlinearities from destroying too much information [17]. The inverted residual block not only significantly reduces the number of parameters through the separable depthwise convolution, but also effectively reserves strong feature extraction capabilities through expanding to a higher dimensional space along with linear bottlenecks.

Additionally, the inverted residual block incorporates the squeeze-and-excitation (SE) block, similar to MobileNetV3 [18]. This block functions as a channel attention mechanism, learning the importance of each channel and subsequently re-weighting each to the more crucial ones. The operation of the SE block, as illustrated in Fig. 1 (b), involves two key steps: squeezing and excitation. During the squeezing phase, the network employs global average pooling across the feature maps of each channel to extract global contextual information, thereby forming a compact channel representation. Following this, in the excitation phase, the SE block uses fully connected layers to learn interdependencies among channels, resulting in a weight vector that reflects the importance of each channel. These weights are then applied to modulate the feature maps of each channel, thus enabling the network to prioritize more significant features for ABP reconstruction. With these enhancements, the encoder of InvResUNet achieves a lightweight structure while maintaining superior feature extraction capabilities.

The decoder of InvResUNet follows the upsampling framework of UNet. The unique skip connection mechanism of U-Net allows for the integration of both low-level fine-grained and high-level contextual features from the encoding stage to the decoder [19]. Additionally, to alleviate the computational demands of the model, this study replaces the traditional 1x3 convolutional kernels in the U-Net upsampling process with depthwise separable convolutions.

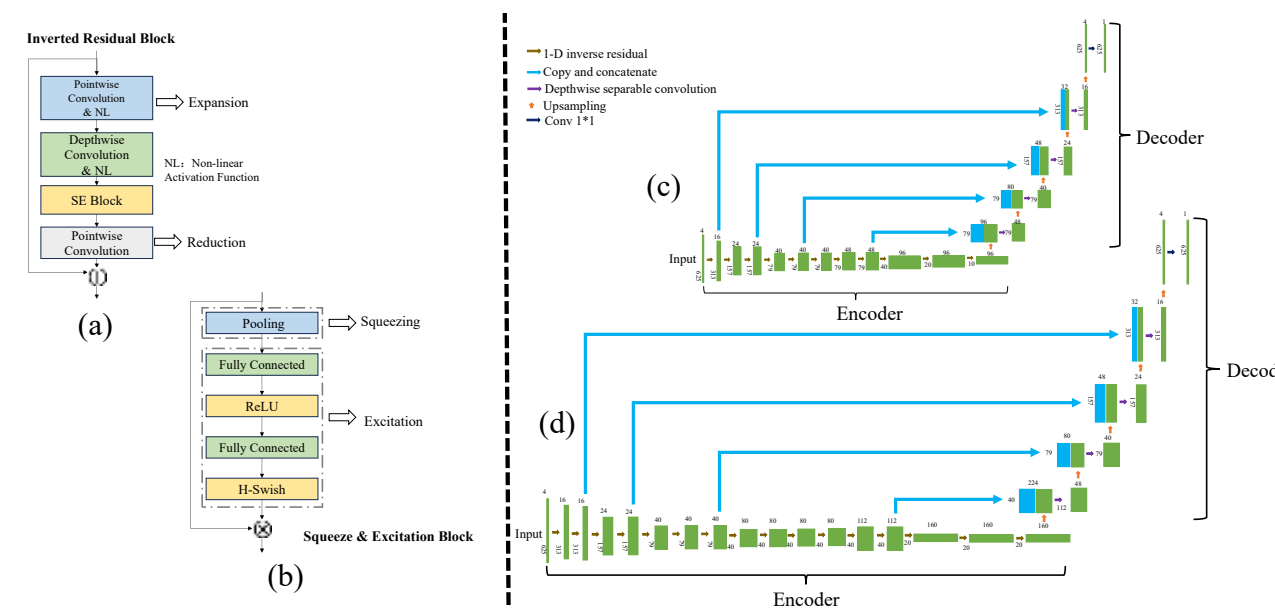

Fig. 1. The architecture of the proposed model InvResUNet. (a) The inverted residual block; (b) the Squeeze-and-Excitation (SE) block; (c) the proposed small InvResUNet network (sInvResUNet); (d) the proposed large InvResUNet network (lInvResUNet).

To determine the most suitable lightweight deep learning network for specific application scenarios, this study expands upon the design of InvResUNet by introducing two structurally similar variants that vary in parameter quantity and network depth. As shown in Fig. 1(c) and (d), the models include sInvResUNet, which has a smaller number of parameters and comprises 11 layers of inverted residual blocks, and lInvResUNet, which has a larger number of parameters and encompasses 16 layers of inverted residual blocks.

### B. Collaborative Training Strategy

One effective way to enhance the performance of a lightweight model is knowledge distillation. In the traditional knowledge distillation framework, the teacher is pre-trained first and then fixed when transferring knowledge to a student, so the knowledge can only be transferred from the teacher to the student. Previous work shows that, a high-quality teacher is important for optimizing a good student [20]. As ABP reconstruction is still a challenging task and a single high-performance teacher model is not available yet, this study proposes to adopt a collaborative learning framework, which makes multiple student models to jointly learn from each other. Specifically, the collaborative framework merges the training processes of all student models with similar or potential different capacities. The diversity in the capacities of students and the mutual learning process enables them to refine their own predictions and develop a more robust understanding of the data, which helps to mitigate overfitting, thus improving overall model performance on unseen data [20].

***Collaborative training framework***. Given that UNet and Transformers have achieved state-of-the-art performance for ABP reconstruction [12], [14], [15], [16], [21], [22], two deep learning models, i.e., UTransBPNet [15], UNet [19], were chosen with the proposed lightweight models, i.e., lInvResUNet and sInvResUNet, to serve as student models for collaborative learning. Specifically, UTransBPNet was selected due to its strong capability in short- and long-range feature representation facilitated by the combined architecture of UNet and Transformer in scenarios with significant intra-subject BP variations.

The training framework, as depicted in Fig. 2, involves a two-phase process. In the first phase, each model was trained independently to allow for effective information extraction on an individual model basis, and the loss of each student model is defined as:

$$L(\hat{y}, y_{gt}) = \frac{1}{N}\sum_{i=1}^{N}\left|y_{gt} - \hat{y}\right|_i \quad (1)$$

where $y_{gt}$ and $\hat{y}$ represent the ground truth and predicted values for each student model, respectively. *N* is the number of samples.

In the second phase, all four models underwent collaborative training, where a second-level representation of their outputs acted as a teacher model. This representation was utilized to adjust the loss function of the student models for knowledge distillation from the teacher model to each student model. Two representative second-level representation were explored in this study: 1) The *minimum* error representation is a straightforward ensembling strategy where the model with the lowest absolute error is selected as the teacher model, as shown in Equation (2); 2) the *average* error strategy emphasizes the robustness of the teacher model by averaging out potential noise from different student models, as shown in Equation (3).

$$\hat{y}_{T,\min} = \hat{y}_k,\ where\ k = \arg\min_j L(\hat{y}_j, y_{gt}) \quad (2)$$

$$\hat{y}_{T,mean} = \frac{1}{M}\sum_{j=1}^{M}\hat{y}_j \quad (3)$$

where M is the number of student models.

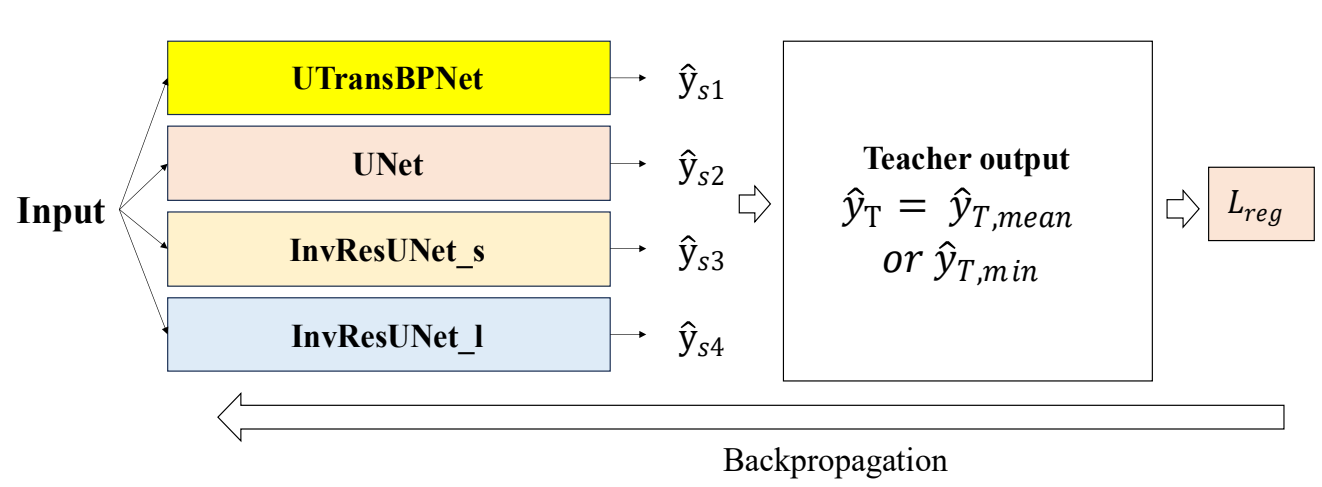

Fig. 2. The proposed collaborative learning framework.

Additionally, recognizing that the teacher model may not always yield accurate predictions, this study employed the Attentive Imitation Loss (AIL) [23] as detailed in Equation (4). This learning loss mechanism uses the teacher loss to re-weight the contributions of the student and the student-teacher imitation losses. Specifically, a larger teacher loss increases the weight of the student loss while decreasing that of the imitation loss and vice versa. This method ensures a dynamic adjustment of learning priorities based on the reliability of the teacher model's output. The AIL loss is defined as:

$$L_{reg} = \frac{1}{N}\sum_{i=1}^{N}\alpha\left|\left|\hat{y}_S - y_{gt}\right|\right|_i^2 + (1-\alpha)\Phi_i\left|\left|\hat{y}_S - \hat{y}_T\right|\right|_i^2 \quad (4)$$

$$\Phi_i = \left(1 - \frac{\left|\left|\hat{y}_T - y_{gt}\right|\right|_i^2}{\eta}\right) \quad (5)$$

$$\eta = max(e_T) - min(e_T) \quad (6)$$

$$e_T = \left\{\left|\left|\hat{y}_T - y_{gt}\right|\right|_i^2;\ i = 1, \dots, N\right\} \quad (7)$$

where $\alpha$ is the hyperparameter to adjust the contribution of the student loss and the student-teacher imitation loss. $\Phi$ represents the factor that dynamically adjusts the contribution of imitation loss according to the teacher loss. $\eta$ applies a min-max normalization to keep $\Phi$ within (0,1).

***Training setup***. The training of the two phases was conducted within a ten-fold subject-independent cross-validation framework. The ten-fold division of the training and testing subsets is consistent across both phases, ensuring there is no risk of data leakage. As the goal of collaborative learning is to train an effective and lightweight model, the training is immediately stopped once no significant improvement was observed on sInvResUNet. This early stopping strategy prevents sInvResUNet from overfitting. The learning rate is set to 1e-5. Training proceeded for a maximum of 100 iterations, with early stopping enabled after 10 iterations if no improvement is observed. The AdamW optimizer is employed, known for its adaptive learning rate and weight decay benefits, set at 0.1 to regularize the model and reduce overfitting. A batch size of 64 is chosen to balance memory efficiency and training speed.

In addition, statistical tests were performed to examine influences of demographic and cardiovascular characteristics on model performance. Specifically, Pearson's correlation analysis was performed between individual MAEs of the proposed model and numerical demographic characteristics including age, height, weight and BMI, as well as cardiovascular parameters represented by heart rate, BP and related variables. Since gender is a categorical characteristic, unpaired Wilcoxon signed-rank test was employed to determine if it had significant influences on model performance by examining if there were significant differences between MAEs of male and female populations. One-way Analysis of Variance (ANOVA) was adopted to further assess the impacts of age on model performance by testing if MAEs and PCCs were significantly different across multiple age groups. Each analysis was performed on 2,154 individuals.

### C. Dataset

The dataset used in this study was derived from a publicly available perioperative dataset curated in [24], which integrates physiological recordings from the VitalDB database. VitalDB contains synchronized electrocardiogram (ECG), photoplethysmogram (PPG), and invasive arterial blood pressure (ABP) signals collected from 2,938 surgical patients at a sampling rate of 125 Hz. Recordings shorter than 20 minutes were excluded, resulting in a total of 2,154 subjects included in this study.

All recordings were segmented into non-overlapping 10-second signal windows. The ECG and PPG signals were downsampled to 62.5 Hz and normalized on a per-segment basis, while ABP signals were preserved in their original scale. In addition to ECG and PPG, the first and second derivatives of the PPG signal were included as model inputs. Segments containing severe artifacts or physiologically implausible values were removed following standard preprocessing procedures described in [24], [25]. After preprocessing, a total of 1,257,141 segments were retained for model development and evaluation. It is featured a heterogenous population with a broad age range from 18 to 97 with mean ± SD of 58.9 ±14.3 years. It also exhibits wide BP ranges (41~257 mmHg for SBP, and 31~234 mmHg for DBP) and substantial intra-subject fluctuations in SBP and DBP. The maximum-to-minimum differences ($\Delta SBP_{max-min}$ and $\Delta DBP_{max-min}$) range from 20 to 180 mmHg, and the mean ± standard deviation for $\Delta SBP_{max-min}$ is 74.2 ± 26.3, and 43.6 ± 20.5 mmHg for $\Delta DBP_{max-min}$.

## III. Results

TABLE I. compared the performance and model complexity of four models with and without collaborative training. Among the four student models, UTransBPNet initially exhibited the highest performance without collaborative training. However, upon incorporating collaborative training, all student models except UTransBPNet showed improved performance, with UNet achieving the most notable enhancement. Notably, sInvResUNet, despite having the smallest model size and lowest computational load, demonstrated comparable accuracy to its larger counterparts. This outcome underscores the effectiveness of the collaborative learning framework in enhancing the capabilities of smaller models. The large version of the proposed lightweight model, lInvResUNet, performed similarly to the smaller version sInvResUNet, but with additional overheads in terms of model size and computational demand. Additionally, the two second-level representation methods also yielded similar performance. Therefore, the remaining analysis primarily focuses on the smaller network, incorporating collaborative learning with mean strategy, referred as KDCL_sInvResUNet. It is noteworthy that the performance of proposed models did not surpass that of the KDCL_UNet model, possibly due to the larger number of parameters and greater network complexities.

TABLE II. presents the ablation results showing the contribution of each module to the performance of the proposed KDCL-sInvResUNet. This includes the removal of the SE module (NoSE), exclusion of one student model (NoUNet and NoUTransBPNet) in the collaborative training, and the omission of the AIL loss function. The results indicate that each module makes a small but meaningful contribution to the model's performance, with the best results achieved when all modules are included in the proposed KDCL-sInvResUNet.

We calculated the Pearson correlation between the estimated and reference ABP to evaluate the capability of four models (UTransBPNet, UNet, sInvResUnet and KDCL_sInvResUNet) in tracking ABP changes. The two lightweight models achieved similar correlation coefficients on individual basis as the large models, i.e., 0.87±0.07 and 0.88±0.07, respectively.

TABLE I. Comparing The Performance, Model Sizes And Computational Load Of The Four Models With And Without Collaborative Training

| Model | MAE, mmHg | STD, mmHg | Number of parameters, M | Computational load, GFLOPs |
|---|---|---|---|---|
| UTransBPNet[15] | 10.24 | 13.32 | 33.19 | 6.36 |
| KDCL-UTransBPNet (Mean) | 11.87 | 15.38 | 33.19 | 6.36 |
| KDCL-UTransBPNet (Min) | 11.54 | 14.90 | 33.19 | 6.36 |
| UNet[19] | 10.31 | 13.41 | 10.82 | 0.85 |
| KDCL-UNet (Mean) | **9.92** | 12.96 | 10.82 | 0.85 |
| KDCL-UNet (Min) | 9.97 | **12.89** | 10.82 | 0.85 |
| sInvResUNet | 10.28 | 13.37 | 0.89 | 0.02 |
| KDCL-sInvResUNet (Mean) | 10.06 | 13.13 | **0.89** | **0.02** |
| KDCL-sInvResUNet (Min) | 10.08 | 13.06 | 0.89 | 0.02 |
| lInvResUNet | 10.27 | 13.47 | 3.02 | 0.05 |
| KDCL-lInvResUNet (Mean) | 10.02 | 13.01 | 3.02 | 0.05 |
| KDCL-lInvResUNet (Min) | 10.01 | 13.02 | 3.02 | 0.05 |

(Note: The prefix "KDCL" refers to the models trained with collaborative learning. The suffix "mean" and "min" represent the form of the second-level representation of a teacher model.)

TABLE II. ABLATION RESULTS TO SHOW THE CONTRIBUTION OF EACH MODULE TO KDCL-SINVRESUNET

| Model | MAE, mmHg | STD, mmHg | Number of parameters, M | Computational load, GFLOPs |
|---|---|---|---|---|
| sInvResUNet (NoSE) | 10.30 | 13.47 | 0.42 | 0.02 |
| sInvResUNet | 10.28 | 13.37 | 0.89 | 0.02 |
| KDCL-sInvResUNet (NoUNet) | 10.19 | 13.25 | 0.89 | 0.02 |
| KDCL-sInvResUNet (NoUTrans) | 10.12 | 13.17 | 0.89 | 0.02 |
| KDCL-sInvResUNet (NoUTrans,NoAIL) | 10.14 | 13.18 | 0.89 | 0.02 |
| KDCL-sInvResUNet | 10.06 | 13.13 | 0.89 | 0.02 |

(For simplification, KDCL-sInvResUNet here denotes sInvResUNet with collaborative training and mean second-level representation.)

### *A. Impacts of Demographics and Cardiovascular Status*

To further analyze the robustness of the proposed model in perioperative settings, we investigated the impact of demographic factors and cardiovascular status on ABP estimation performance across different subject subgroups.

***Impact of demographic factors.*** Subjects were grouped according to age, BMI, and gender, and one-way ANOVA tests were performed to assess performance differences, as shown in Fig. 3.As illustrated in Fig. 3(a), all models exhibited statistically significant performance differences across age groups ($p < 0.05$), with higher MAEs observed in elderly subjects (69–90 years). The performance variation across age groups was approximately 1 mmHg. BMI also showed a moderate influence on model performance (Fig. 3 (b)), where overweight subjects (BMI > 25 kg/m²) demonstrated slightly lower MAEs than the normal-weight group. In contrast, gender did not have a statistically significant effect on ABP estimation accuracy for any model (Fig. 3 (c)).

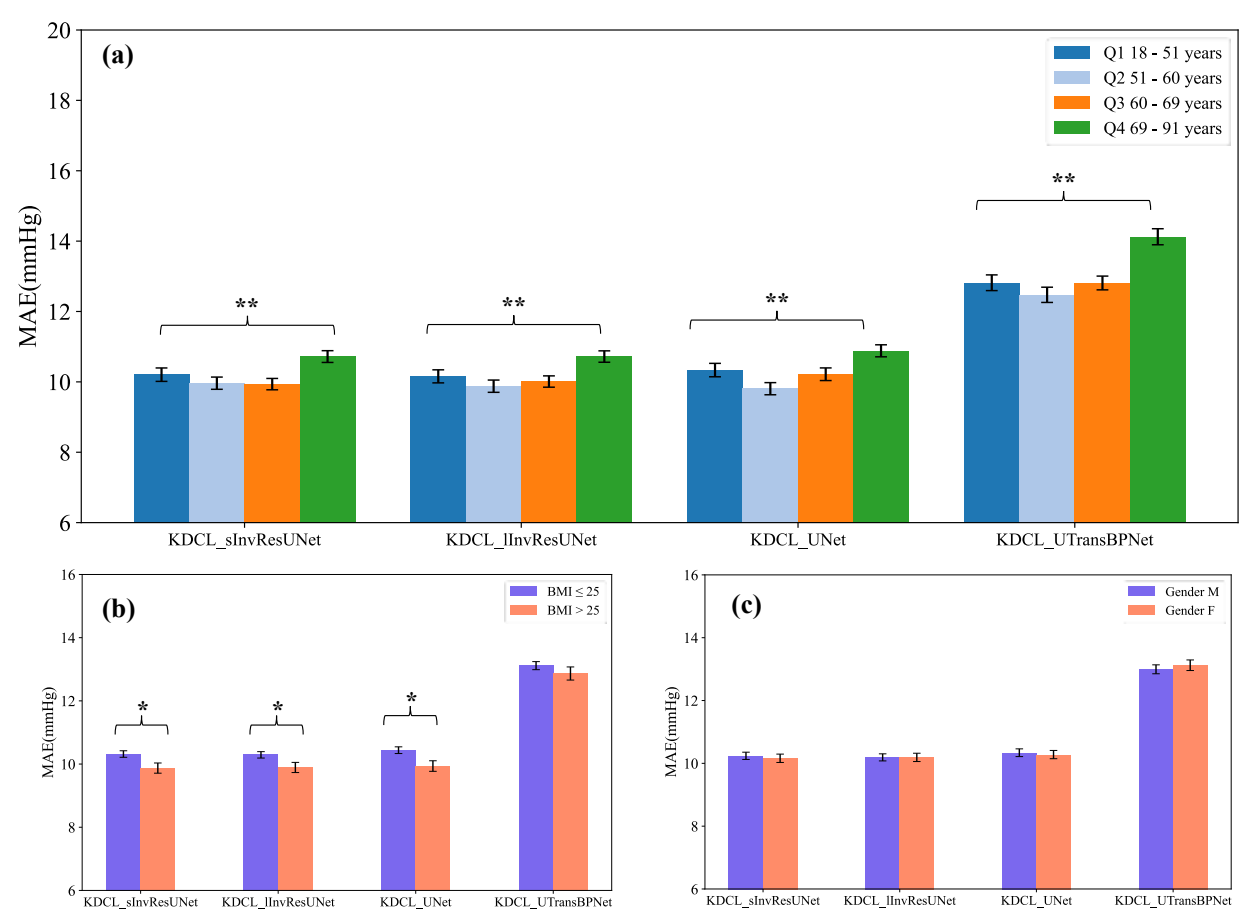

Fig. 3. MAEs of different demographic groups for the four models. (a) Age. (b) BMI. (c) Gender. One-way ANOVA tests were performed across demographic groups. * indicates $p < 0.05$, ** indicates $p < 0.01$. First quantile (Q1): from the minimum to the 25th percentile of the data; Q2: from the 25th percentile to the median; Q3: from the median to the 75th percentile; Q4: from the 75th percentile to the maximum.

***Impact of cardiovascular status.*** The influence of cardiovascular characteristics was further examined by dividing subjects into quantiles based on each variable. The results for representative SBP-related variables are presented in Fig. 4. All cardiovascular variables in Fig. 4 showed statistically significant effects on model performance ($p < 0.05$). Static BP indicators, such as $SBP_{mean}$ and $SBP_{max}$, exhibited a center-low-and-extreme-high error pattern, whereas dynamic indicators, including $SBP_{std}$ and $\Delta SBP_{max-min}$, showed a clear positive correlation with MAEs. In particular, larger intra-subject BP fluctuations were consistently associated with increased estimation errors. Among all variables, $SBP_{mean}$ had the most pronounced impact, with MAE differences exceeding 2 mmHg across quantiles.

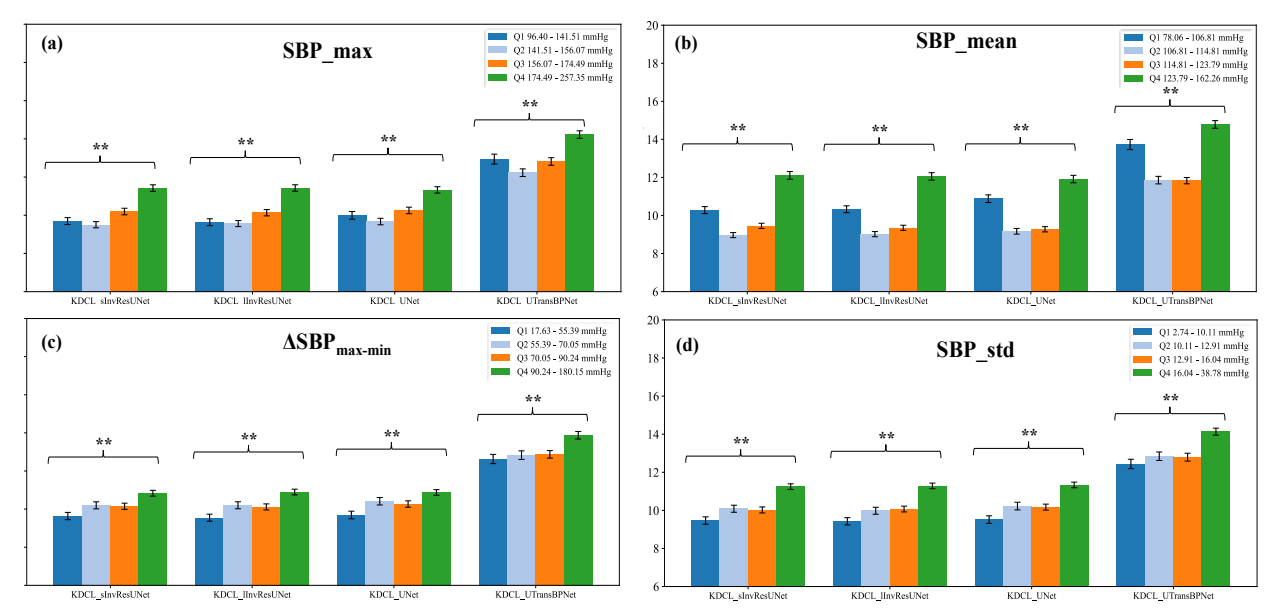

Fig. 4. MAEs at different quantiles of cardiovascular variables for the four models. (a) $SBP_{max}$. (b) $SBP_{mean}$. (c) $\Delta SBP_{max-min}$. (d) $SBP_{std}$. First quantile (Q1): from the minimum to the 25th percentile of the data; Q2: from the 25th percentile to the median; Q3: from the median to the 75th percentile; Q4: from the 75th percentile to the maximum.

***Case analysis.*** As shown in Fig. 5, for subject (a), when the intra-subject BP level is close to the population average ($SBP_{mean}$ = 114.5 mmHg，where $SBP_{mean}$ denotes the mean systolic blood pressure of the subject) and the intra-subject fluctuation and range are relatively low ($SBP_{std}$ = 4.7 mmHg, $SBP_{max}$ = 128.2 mmHg, where $SBP_{std}$ and $SBP_{max}$ denote the standard deviation and maximum value of SBP, respectively), the four models can effectively predict the ABP waveform, with the KDCL_sInvResUNet predicted the closest to the reference waveform. However, in subject (b) with larger deviations from the population average and higher fluctuation and range ($SBP_{mean}$ = 130.7 mmHg, $SBP_{std}$ = 8.75 mmHg, $SBP_{max}$= 155.7 mmHg), all models failed to accurately track the reference waveform. Moreover, it was observed that UTransBPNet showed notable jitter in its predictions.

### *B. Embedded Deployment*

In this section, KDCL_sInvResUNet was deployed on embedded systems to test the feasibility for real-time inference of ABP. Two embedded platforms were evaluated, i.e., the widely used Raspberry Pi 4 Model B and the NVIDIA

Jetson TX2 NX development board. ECG and PPG signals were acquired in real time by the Biopac MP160 vital signal acquisition device, and transmitted to an embedded platform via the TCP/IP protocol. Subsequently, the signals were preprocessed on these embedded platforms, involving downsampling, segmentation and normalization, to ensure compatibility with the input of KDCL_sInvResUNet. The preprocessed data was then fed into the ONNX Runtime environment, which was loaded with the KDCL_sInvResUNet model for real-time ABP estimation. The deployment process is illustrated in Fig. 6. **错误!未找到引用源。** shows the inference time of a 10-second signal segment of KDCL_sInvResUNet on the two embedded systems. The result suggests that KDCL_sInvResUNet can achieve real-time monitoring when deployed on both boards.

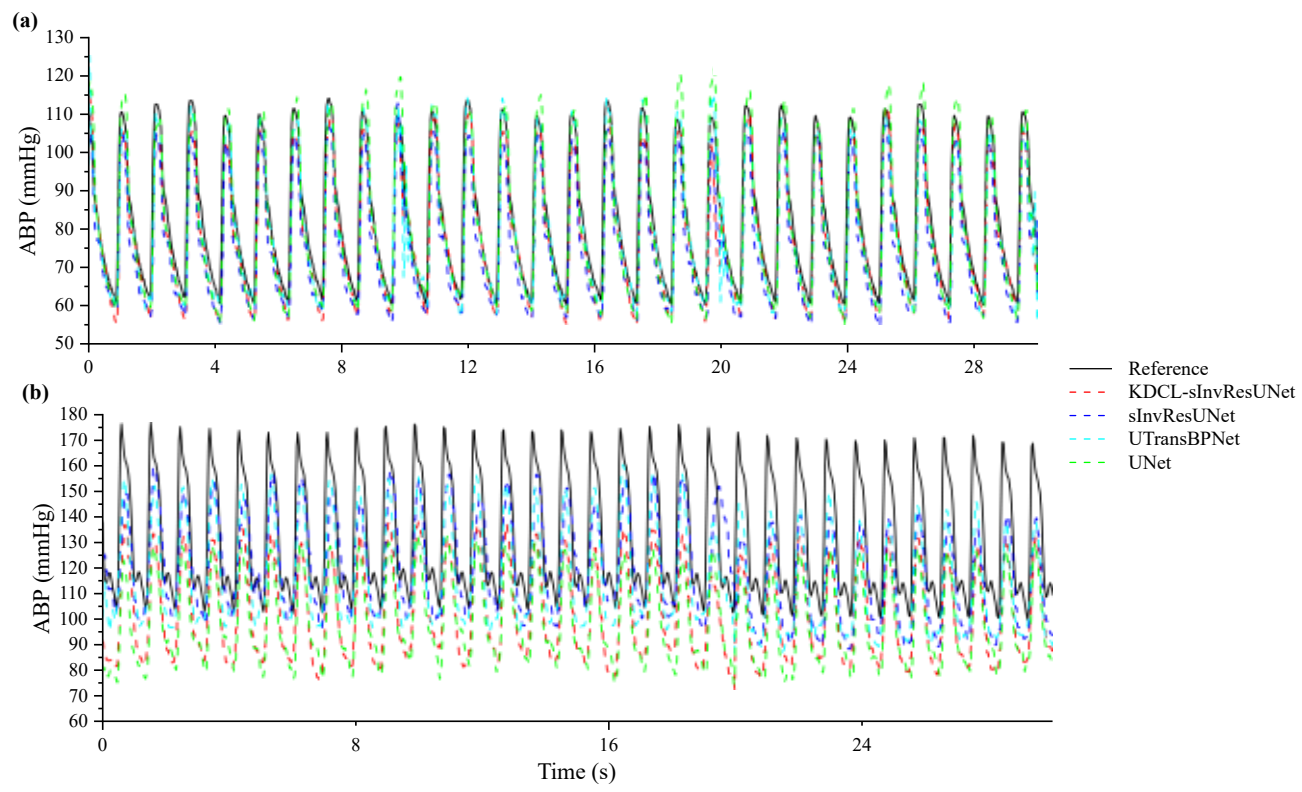

Fig. 5. Case analysis of ABP reconstruction by the four models. (a) One subject with low intra-subject BP level, fluctuation and range ($SBP_{mean}$ = 114.5 mmHg, $SBP_{std}$ = 4.7 mmHg, $SBP_{max}$ = 128.2 mmHg); (b) another subject with high intra-subject BP level, fluctuation and range ($SBP_{mean}$ = 134.6 mmHg, $SBP_{std}$ = 14.8 mmHg, $SBP_{max}$ = 174.1 mmHg).

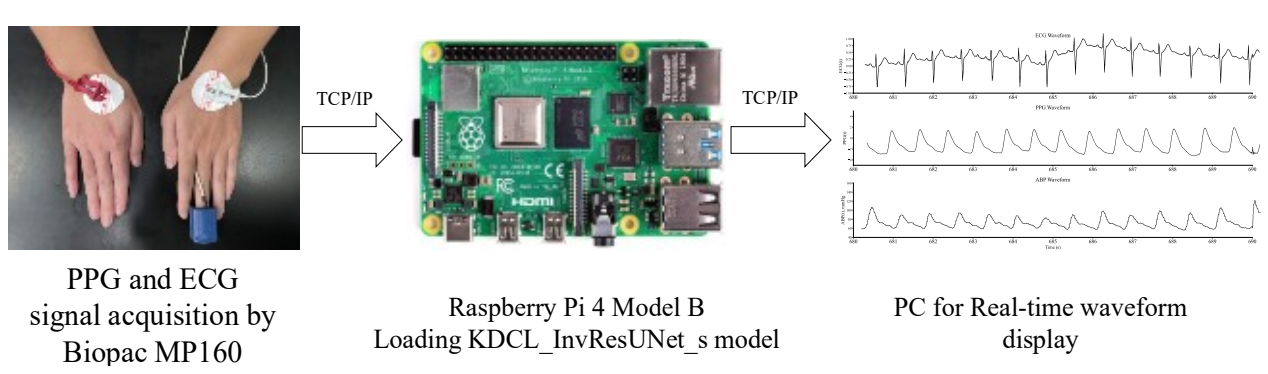

Fig. 6. Flowchart of deployment of the KDCL_sInvResUNet model on an embedded device.

TABLE III. INFERENCE TIME FOR A 10-SECOND SEGMENT OF KDCL_SINVRESUNET ON THE TWO EMBEDDED DEVICES

| Development board | Inference time |
|---|---|
| Raspberry Pi 4 Model B | 8.49 ms |
| NVIDIA Jetson TX2 NX | 6.95 ms |

## IV. DISCUSSION

This study presents a lightweight deep learning framework for continuous ABP waveform reconstruction from ECG and PPG signals, with a primary focus on embedded deployment. By integrating inverted residual blocks and depthwise separable convolutions into a one-dimensional U-Net architecture together with the introduction of the collaborative training, the proposed KDCL_sInvResUNet achieves a favorable balance between reconstruction accuracy and computational efficiency.

Compared with existing lightweight ABP reconstruction methods, the proposed KDCL_sInvResUNet demonstrates clear advantages in computational efficiency and inference latency for embedded deployment. BPNet[16] reported an inference time of 42.53 ms on a Raspberry Pi 4 Model B, which is approximately 5 times slower than our implementation (8.49 ms), while its model size was not disclosed. KD-Informer[14] achieved a comparable parameter size (0.81 M vs. 0.89 M), but required substantially higher computational cost (0.19 GFLOPs vs. 0.02 GFLOPs). Other lightweight models, such as DiffCNBP[26], rely on considerably larger memory footprints and computational resources, limiting their practicality for real-time edge deployment.

It is worth noting that, all models in our study struggled to achieve performance required by the IEEE Standard 1708 which necessitates the MAE lower than 7 mmHg. This may be attributed to the challenging nature of this dataset, which spans a diverse demographic population and covers a wide range of BP.

Furthermore, our statistical analysis indicates that demographic factors, particularly age and BMI, systematically influence the performance of all evaluated models, including complex architectures such as UNet and UTransBPNet. Across models, estimation accuracy was highest for subjects whose age and BP values were close to the population mean, while performance degraded at the extremes of age and BP. This trend is consistent with the imbalanced distribution of demographic characteristics and BP values in the dataset, suggesting that current population-based models are biased toward majority groups and struggle to generalize to underrepresented subpopulations.

Moreover, population-based training complicates the physiological interpretation of BP values. In perioperative settings, interventions such as fluid administration and vasopressor use can induce BP changes that deviate from an individual's baseline to varying extents. These deviations may activate different regulatory mechanisms (e.g., baroreflex, renin–angiotensin system, endothelial responses), which can operate independently or synergistically depending on the magnitude of BP change. As a result, identical BP values may reflect different physiological states across individuals.

## V. CONCLUSION

This study proposed a lightweight deep learning model, KDCL_sInvResUNet, for real-time ABP estimation on resource-constrained embedded devices. By validating on a real-world perioperative dataset that encompasses diverse demographic characteristics and dynamic cardiovascular conditions, the research demonstrates KDCL_sInvResUNet achieves comparable accuracy to large models, but with significantly reduced model size and computational load, making it feasible for real-time deployment on embedded systems. Through a comprehensive analysis of the factors that influence model performance, we identified significant performance biases across certain demographic groups and cardiovascular conditions for current population-based deep learning models. Our study highlights the need for models that can better handle such diversity.

*Availability of the code.* This study has open-sourced the code of the proposed model: https://github.com/SZTU-wearable/KDCL_InvResUNet.